\documentclass[10pt,twocolumn,letterpaper]{article}

\usepackage[pagenumbers]{cvpr} 
\usepackage{cvpr}              

\definecolor{cvprblue}{rgb}{0.21,0.49,0.74}
\usepackage[pagebackref,breaklinks,colorlinks,allcolors=cvprblue]{hyperref}

\usepackage{booktabs}
\usepackage{algorithm}
\usepackage{algorithmic}
\usepackage{multirow}
\usepackage{gensymb}
\usepackage{makecell}
\usepackage{adjustbox}
\usepackage{threeparttable}
\usepackage{bbding}
\usepackage{makecell,pifont,multirow,multicol}
\usepackage{colortbl}
\usepackage[misc]{ifsym}

\def\paperID{4900} 
\def\confName{CVPR}
\def\confYear{2025}

\usepackage{hyperref}

\usepackage{orcidlink}

\def\ModelName{UniMamba}

\def\logo{\makebox[22pt][l]{\raisebox{-0.9ex}{\includegraphics[height=25pt]{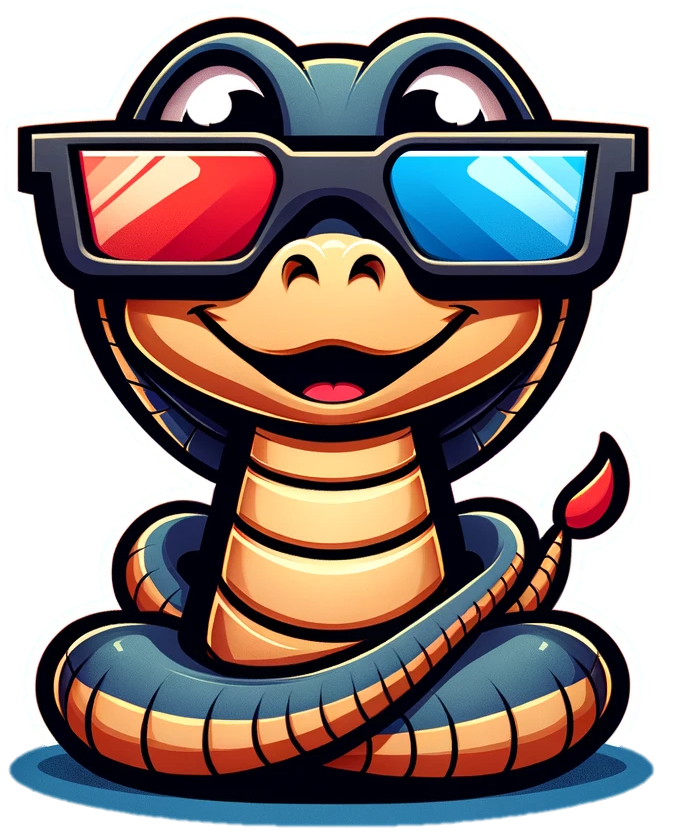}}}}

\title{\logo\ModelName: Unified Spatial-Channel Representation Learning with Group-Efficient Mamba for LiDAR-based 3D Object Detection}

\author{Xin Jin$^{\spadesuit2,3\heartsuit}$ \quad
Haisheng Su$^{\spadesuit1,3}$$^{\textrm{\Letter}}$ \quad
Kai Liu$^{3}$ \quad
Cong Ma$^{3}$ \\
Wei Wu$^{3,4}$ \quad
Fei HUI$^{2}$$^{\textrm{\Letter}}$ \quad
Junchi Yan$^{1}$$^{\textrm{\Letter}}$ \\
$^{1}$School of Computer Science, Shanghai Jiao Tong University \\ 
$^{2}$Chang'an University, $^{3}$SenseAuto Research, $^{4}$Tsinghua University \\
{\tt\small \{suhaisheng,yanjunchi\}@sjtu.edu.cn, \{jinxin,feihui\}@chd.edu.cn} \\
{\tt\small \{liukai3.iag,macong,wuwei\}@senseauto.com} \\
\vspace{-0.6cm}
}

\begin{document}

\maketitle

\let\thefootnote\relax\footnotetext{$\spadesuit$ Equal Contribution.\quad \quad $\textrm{\Letter}$ Corresponding Authors.}
\let\thefootnote\relax\footnotetext{$\heartsuit$ Intern at SenseAuto Research.}

\begin{abstract}

Recent advances in LiDAR 3D detection have demonstrated the effectiveness of Transformer-based frameworks in capturing the global dependencies from point cloud spaces, which serialize the 3D voxels into the flattened 1D sequence for iterative self-attention. However, the spatial structure of 3D voxels will be inevitably destroyed during the serialization process. Besides, due to the considerable number of 3D voxels and quadratic complexity of Transformers, multiple sequences are grouped before feeding to Transformers, leading to a limited receptive field. Inspired by the impressive performance of State Space Models (SSM), in this paper, we propose a novel Unified Mamba (UniMamba), which seamlessly integrates the merits of 3D convolution and SSM in a concise multi-head manner, aiming to perform ``local and global" spatial context aggregation efficiently and simultaneously. Specifically, a UniMamba block is designed which mainly consists of spatial locality modeling, complementary Z-order serialization and local-global sequential aggregator. The spatial locality modeling module integrates 3D submanifold convolution to capture the dynamic spatial position embedding before serialization. Then the efficient Z-order curve is adopted for serialization both horizontally and vertically. Furthermore, the local-global sequential aggregator adopts the channel grouping strategy to efficiently encode both ``local and global" spatial inter-dependencies using multi-head SSM. Additionally, an encoder-decoder architecture with stacked UniMamba blocks is formed to facilitate multi-scale spatial learning hierarchically. Extensive experiments are conducted on three popular datasets: nuScenes, Waymo and Argoverse 2. Particularly, our UniMamba achieves 70.2 mAP on the nuScenes dataset.

\end{abstract}    
\vspace{-0.2cm}
\section{Introduction}
\label{sec:intro}

\begin{figure}
    \centering
    \includegraphics[width=0.95\linewidth]{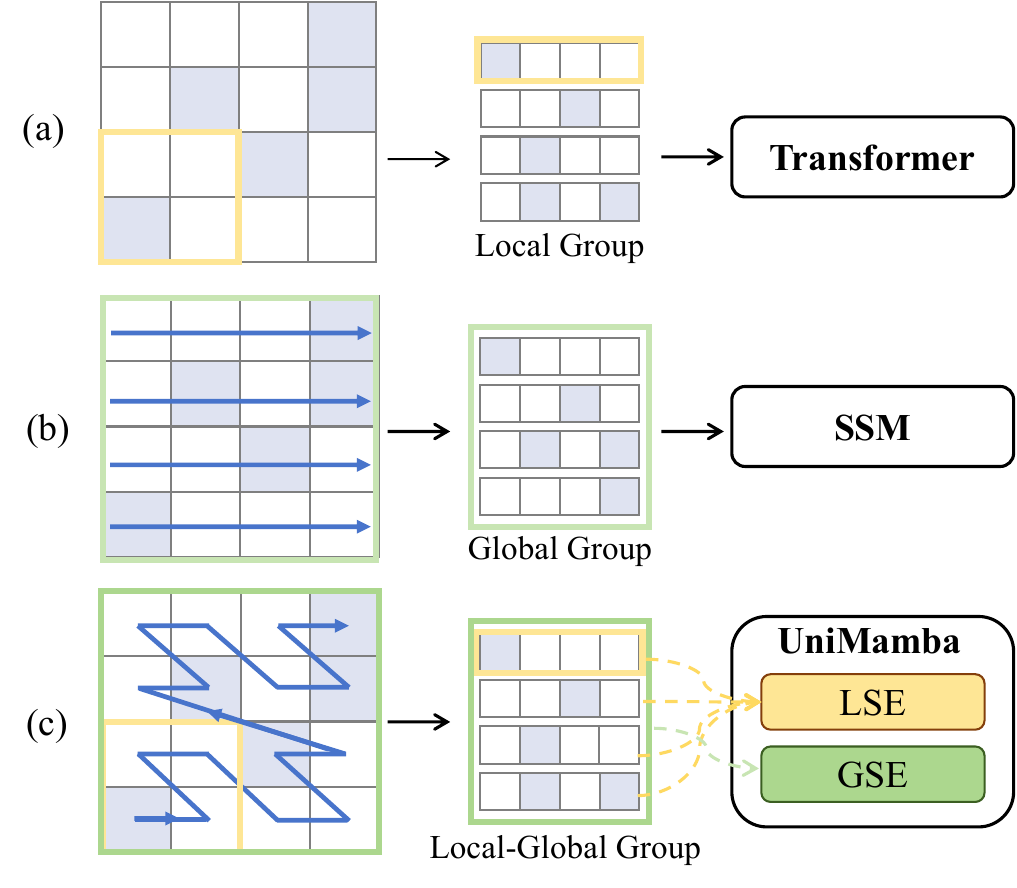}
    \vspace{-0.4cm}
    \caption{Comparison of different 3D backbones. (a) Transformer-based backbone using \textbf{local} window grouping. (b) SSM-based backbone using \textbf{global} sequence grouping. (c) Our proposed UniMamba backbone using channel-wise \textbf{local-global} grouping.} 
    \label{fig:1}
    \vspace{-0.3cm}    
\end{figure}




LiDAR is widely used in Autonomous Driving perception tasks~\cite{su2024robosense,su2024difsd} due to its ability to provide precise 3D location information. However, unlike structured image data, the sparsity and disorder of point cloud space pose significant challenges for accurate 3D object detection. Therefore, learning effective pattern representations from 3D point clouds is a key focus of current research.

Previous methods are generally classified into point-based and voxel-based approaches. Point-based methods~\cite{shi2019pointrcnn,yang20203dssd} directly apply operators such as PointNet~\cite{qi2017pointnet,qi2017pointnet++} to manipulate raw point clouds, which entails a time-consuming process of sampling and grouping, leading to low computational efficiency. In contrast, voxel-based methods~\cite{zhou2018voxelnet,yan2018second,chen2023largekernel3d} convert point clouds into regular grids, showcasing superior performance and becoming the dominant mainstream architecture. 

Voxel-based methods can be further categorized into sparse convolution neural networks (SpCNN)-based and Transformer-based. SpCNN-based approaches~\cite{zhang2024safdnet,yin2021center} are constrained by the inherently small receptive field of convolutions, which limits their ability to extract rich contextual information. Transformer-based methods~\cite{mao2021voxel,liu2023flatformer, wang2023dsvt} suffer from quadratic complexity, which tend to serialize 3D voxels into multiple 1D sequences and adopt self-attention mechanism respectively, leading to limited receptive field by group size and feature size (Fig.~\ref{fig:1} (a)). However, both local structure details and global context are crucial \cite{li2022uniformer, li2022uniformerv2,qing2021temporal} for 3D object detection from sparse and occluded point clouds. \textit{Under this circumstance, a robust LiDAR 3D backbone with flexible receptive fields and great spatial modeling capacity as well as computational efficiency is worth exploring.}


Inspired by the remarkable success of the Mamba~\cite{mamba, vim,liu2024lion,liu2024vmamba} architecture in the areas of vision tasks, we investigate its applicability to LiDAR 3D detection. Its linear complexity enables the whole sequence processing without voxel grouping (Fig.~\ref{fig:1} (b)), thus facilitating the global context modeling \cite{zhang2024voxel}. However, simple adaptation of Mamba to LiDAR 3D backbone may have some \textit{drawbacks}: \textbf{(1)} loss of spatial voxel locality during serialization; \textbf{(2)} local redundancy brought by coherent global modeling, causing inferior efficiency; \textbf{(3)} lack of spatial diversity to handle complex local and global dependencies.


To this end, we propose \textbf{UniMamba}, a unified Mamba architecture to integrate 3D convolution and State Space Models (SSM) in a concise multi-head format, which can achieve a preferable balance between effectiveness and efficiency in spatial structure modeling for LiDAR 3D object detection. Specifically, our UniMamba consists of three main modules, namely spatial locality modeling, complementary Z-order serialization and local-global sequential aggregator. The Spatial Locality Modeling (SLM) module aims to capture the dynamic structure embedding with 3D submanifold sparse convolution before serialization. Then the complementary Z-order serialization is proposed to transform 3D voxels into 1D sequence, considering the spatial proximity preservation both horizontally and vertically. Subsequently, a Local-Global Sequential Aggregator (LGSA) is designed to perform ``local-global" spatial aggregation in a group-efficient fashion (Fig.~\ref{fig:1} (c)), which includes two sub-modules, \textit{i.e.,} Local Sequential Encoder (LSE) and Global Sequential Encoder (GSE). Concretely, GSE processes all voxels in a single 1D sequence with multi-head bidirectional SSM, allowing global contextual modeling directly without multiple sub-sequence grouping. Contrary, LSE handles the grouped 1D sequences respectively to capture the local inter-dependencies. Besides, a channel grouping strategy is proposed to aggregate ``local-global" spatial context in parallel. Finally, through stacking our UniMamba blocks into an encoder-decoder architecture with several stages, our UniMamba exhibits great potential as LiDAR 3D backbone for multi-scale spatial relation modeling efficiently, achieving 70.2 mAP and 74.0 NDS on nuScenes test set particularly. In summary, the main contributions of our work can be summarized into three folds:


\begin{itemize}
    \item We propose \textbf{UniMamba}, a novel unified 3D backbone for LiDAR 3D object detection, which integrates 3D convolution and bidirectional SSM to achieve effective spatial modeling in a group-efficient manner.
    \item We design a Local-Global Sequential Aggregator to simultaneously capture \textit{local and global} voxel relations in a channel-grouping fashion, benefiting from the spatial locality modeling and complementary Z-order serialization in case of spatial proximity loss.
    \item Extensive experiments are conducted on three popular benchmarks, \textit{i.e.,} nuScenes, Waymo and Argoverse 2, demonstrating the effectiveness of our UniMamba.
\end{itemize}

\section{Related Work}
\label{sec:relatedwork}

\begin{figure*}
    \centering
    \includegraphics[width=\linewidth]{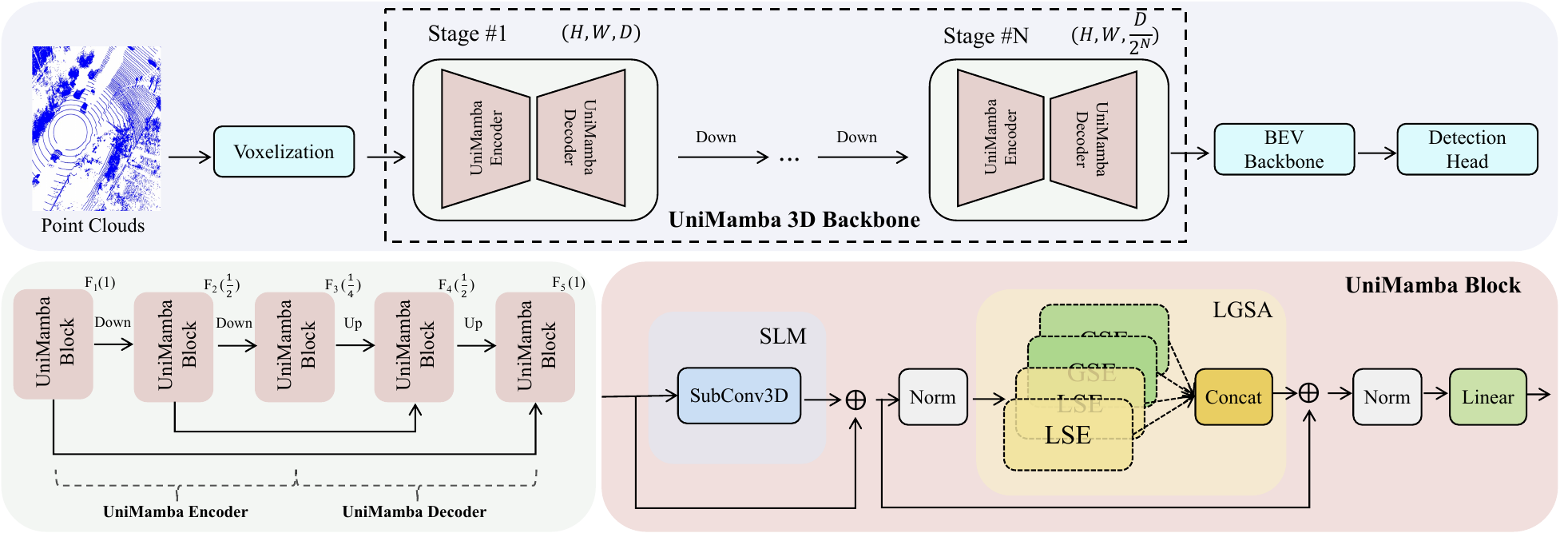}
    \vspace{-0.5cm}
    \caption{Illustration of our proposed UniMamba backbone, which consists of multiple stages, and each stage includes several UniMamba blocks to encode multi-scale features with an encoder-decoder architecture through down/up-sampling and stacking operations. The UniMamba Block is our core component, which efficiently enables simultaneous extraction and aggregation of \textit{local and global} contextual information. In UniMamba, We first voxelize the point clouds, then adopt the proposed UniMamba 3D backbone to extract multi-scale rich spatial contextual features. Finally, these enhanced features are fed into a BEV backbone and a detection head for final 3D object detection.}
    \vspace{-0.2cm}
    \label{fig:framework}
\end{figure*}

Lidar-based 3D detectors are typically classified into point-based and voxel-based. Point-based methods~\cite{qi2017pointnet,qi2017pointnet++,shi2019pointrcnn,yang20203dssd,zhang2022not} extract multi-scale local contextual features directly from downsampled raw points. However, the downsampling and multi-scale grouping lead to a significant computational overhead, resulting in low inference speeds that limit their practical applications.  Voxel-based detectors~\cite{zhou2018voxelnet, yan2018second} convert unordered point clouds into structured grids through voxelization and apply various backbones to extract 3D features, which have become mainstream frameworks for outdoor 3D detection tasks. According to the network architecture, 3D backbones can be classified into the following three categories.

\subsection{3D Object Detection with Sparse Convolution}

VoxelNet~\cite{zhou2018voxelnet} pioneered voxel-based detectors by directly utilizing regular convolution. However, due to the sparsity of point clouds, the direct application of 3D convolutions incurs significant computational costs and information redundancy. Therefore, existing methods~\cite{bai2022transfusion,chen2022focal,chen2023voxelnext,deng2021voxelrcnn,zhang2024safdnet} typically use sparse convolutional neural network (SpCNN) to extract features. Largekernel3D~\cite{chen2023largekernel3d} migrate large convolutional kernels into the 3D detection domain, enhancing the inherent receptive field. HEDNet~\cite{zhang2023hednet} learns long-range dependencies via a multi-scale encoder-decoder structure. These methods aim to improve the receptive field of SpCNN-based 3D backbones. However, the advantage of convolution lies in its powerful local information extraction capabilities and inductive bias. Therefore, in this paper, we combine SpCNN to compensate for the lack of locality in mamba when extracting long-range contextual information.

\subsection{3D Object Detection with Transformer}

Transformer~\cite{vaswani2017attention} has been extensively researched in point cloud detection tasks. Due to computational efficiency constraints, it is impractical to directly compute the attention scores for all voxels. Existing methods~\cite{fan2022embracing,zhou2023octr,yang2023pvt,wang2023dsvt,sun2022swformer,liu2023flatformer} typically employ a window-based attention mechanism, dividing the 3D space into multiple small windows, and then applying shift window or multi-scale fusion to achieve a global receptive field. However, small windows still limit the receptive field. In this paper, we utilize Mamba to replace Transformer, achieving larger window groupings, which addresses the issue of limited receptive fields while also providing better inference efficiency.

\subsection{Vision Task with Mamba}

Mamba~\cite{mamba} is a popular sequential modeling architecture with linear complexity. Due to its great computational efficiency, it has been regarded as a strong competitor to Transformer. Some research has begun to explore its potential in vision tasks~\cite{liu2024vmamba,huang2024localmamba,li2024videomamba,han2024mamba3d,liang2024pointmamba}. 
As a sequence model, most existing studies have focused on exploring different scanning methods. 
Vim~\cite{vim} introduced a bidirectional SSM to learn image features. GrootVL~\cite{xiao2024grootvl} adopted a tree topology to construct scanning sequences, enhancing long-range interactions. The most advanced lidar detectors utilized the Mamba architecture, while LION \cite{liu2024lion} developed a unified RNN framework using group-based strategy. VoxelMamba \cite{zhang2024voxel} introduced a group-free approach to directly model global information. In this paper, we investigate the feasibility of Mamba-style architecture for LiDAR 3D detection, and propose UniMamba, which can effectively capture local and global voxel interactions simultaneously.



\section{Method}
\label{sec:method}

\begin{figure*}
    \centering
    \includegraphics[width=\linewidth]{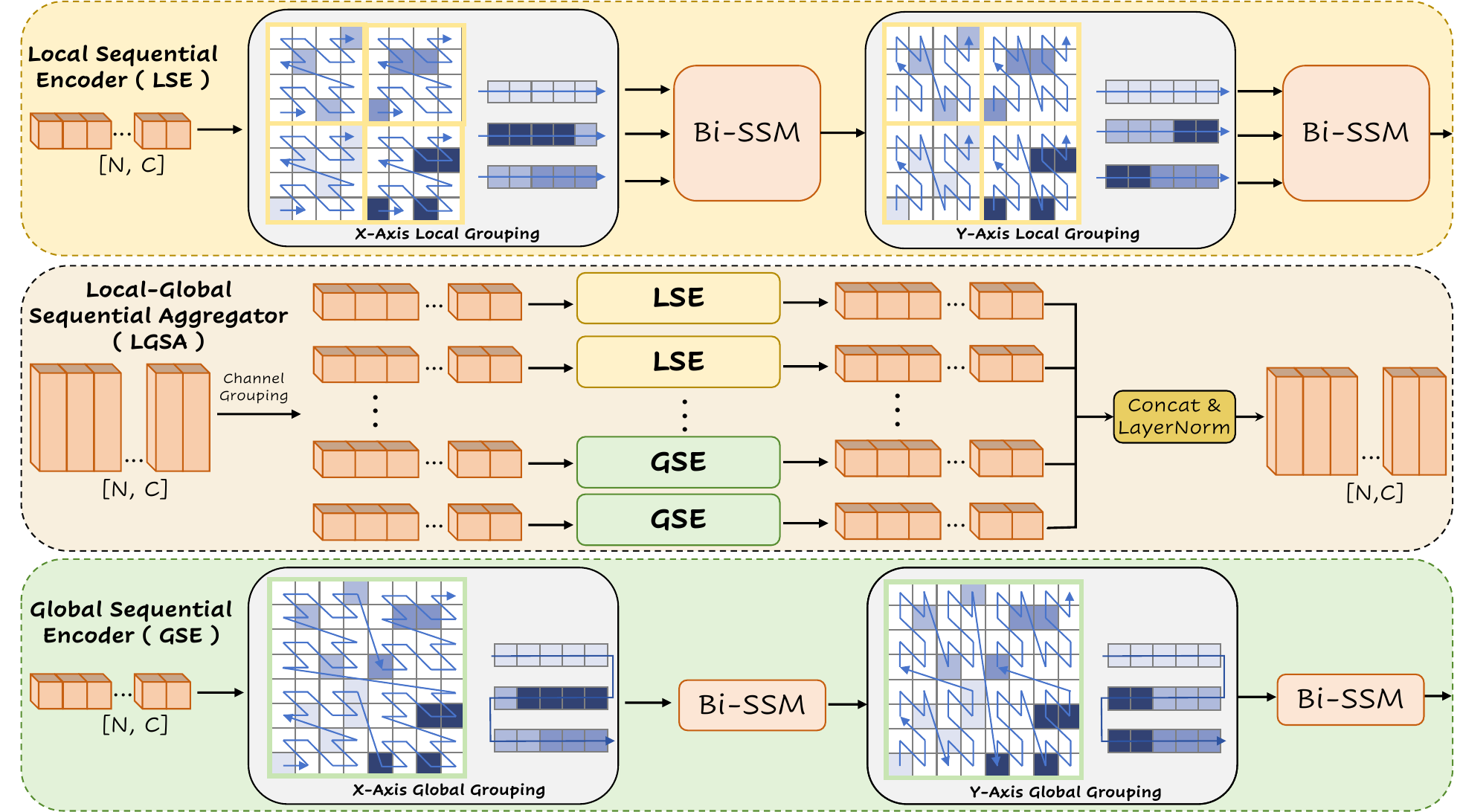}
    \vspace{-0.5cm}
    \caption{Illustration of Local-Global Sequential Aggregator. Local Sequential Encoder (LSE) adopts the bidirectional SSM to handle the multiple 1D groups respectively with the proposed complementary Z-order serialization both vertically and horizontally. Instead, Global Sequential Encoder (GSE) handles a single group without sequence partition to capture the global inter-dependencies. Then the Local-Global Sequential Aggregator (LGSA) combines these two encoders in a multi-head format through channel grouping, which can model both local structure details and global context information simultaneously.}
    \label{fig:lgsa}
    \vspace{-0.2cm}
\end{figure*}

In this section, we introduce the specific architecture of UniMamba, which is a voxel-based 3D backbone integrating 3D convolution and bidirectional SSM to achieve effective spatial modeling in a group-efficient manner. First, we present the basic concepts of Mamba (Sec.~\ref{mamba}), followed by an overview of UniMamba (Sec. \ref{overall}). Then we describe the key components of the UniMamba block (Sec. \ref{unimamba}) in detail. Finally, we provide a detailed analysis of the overall composition of our backbone (Sec. \ref{backbone}).

\subsection{Preliminaries}
\label{mamba}
Mamba is a powerful discrete variant of the State Space Model (SSM). The SSM is a continuous system that establishes a mapping between inputs $x(t) \in \mathbb{R}^{L}$ and output $y(t) \in \mathbb{R}^{L}$ through a hidden state vector $h(t) \in \mathbb{R}^{N}$. This system can be expressed as follows:

\vspace{-0.1cm}
\begin{equation}
    \begin{aligned}
h^{\prime}(t) & =\mathbf{A} h(t)+\mathbf{B} x(t), \\
y(t) & =\mathbf{C} h(t),
\end{aligned}
\end{equation}
where $\textbf{A} \in \mathbb{R}^{N \times N}$, $\textbf{B} \in \mathbb{R}^{N \times 1}$ and $\textbf{c} \in \mathbb{R}^{1 \times N}$ represent the learnable evolution parameters and the two projection parameters, respectively. 
To discretize it for fitting sequences and image data, the zero-order hold (ZOH)method is used to convert $\textbf{A}$ and $\textbf{B}$ into discrete parameters $\overline{\mathbf{A}}$ and $\overline{\mathbf{B}}$ with a time scale parameter $\Delta$. The conversion process is as follows:
\begin{equation}
    \begin{aligned}
&\overline{\mathbf{A}} =\exp(\Delta\mathbf{A}), \\
&\overline{\mathbf{B}} =(\mathbf{\Delta A})^{-1}(\exp(\mathbf{\Delta A})-\mathbf{I})\cdot\mathbf{\Delta B} 
\end{aligned}
\end{equation}

Compared to previous SSMs, Mamba enhances contextual awareness by introducing a dynamic selective scanning mechanism (S6). Specifically, its parameters $\textbf{B} \in \mathbb{R}^{B \times L \times N}$, $\textbf{C} \in \mathbb{R}^{B \times L \times N}$ and $\mathbf{\Delta} \in \mathbb{R}^{B \times L \times D}$ are directly obtained from the input $x \in \mathbb{R}^{B \times L \times D}$. Thus, the discretized SSM can be expressed as:
\vspace{-0.1cm}
\begin{equation}
    \begin{aligned}&h_{t}=\overline{\mathbf{A}}h_{t-1}+\overline{\mathbf{B}}x_{t},\\&y_{t}=\mathbf{C}h_{t}.\end{aligned}
\end{equation}

\subsection{Overview}
\label{overall}
As shown in Fig.~\ref{fig:framework}, UniMamba comprises four main components: the Voxel Feature Encoder,  a 3D backbone, a BEV backbone, and a detection head. Our contribution lies in proposing a novel 3D backbone designed to efficiently capture flexible contextual information. The key component is the UniMamba Block, which first employs a spatial locality modeling module (SLM) (Sec. \ref{slm}) to capture the dynamic structure embedding, and then utilizes a complementary Z-order (Sec. \ref{zorder}) to convert 3D voxels into a 1D sequence. To simultaneously capture contextual information from different receptive fields, we use Local-Global Sequential Aggregator (LGSA) (Sec. \ref{lgme}) to encode both local and global sequences. Additionally, similar to~\cite{zhang2023hednet}, we adopt an encoder-decoder structure to stack UniMamba blocks, forming the final backbone architecture (Sec. \ref{backbone}).

\subsection{UniMamba Block}
\label{unimamba}

\subsubsection{Spatial Locality Modeling}
\label{slm}

Like other Mamba-based architectures~\cite{liu2024vmamba, huang2024localmamba,han2024mamba3d, zhang2024voxel}, compressing 2D/3D space into a 1D sequence inevitably results in the loss of local spatial position information. To address this issue, most methods often design various complex serialization techniques to maintain locality. However, these approaches tend to incur high computational costs and yield only limited effectiveness. Inspired by previous works~\cite{liu2024lion,wu2023point} that use sparse convolutions to provide positional information, we introduce a simple submanifold convolution as Spatial Locality Modeling module (SLM) within the UniMamba block:
\vspace{-0.1cm}
\begin{equation}
    SLM(x) = SubConv3D(x)
\end{equation}
Here, $SubConv3D(\cdot)$ refers to a standard 3D submanifold convolution. Thanks to the manifold design, it possesses geometric adaptability, effectively capturing local features and maintaining shape information. Therefore, we utilize it to compensate for losing local positional information.

\subsubsection{Complementary Z-order Serialization} 
\label{zorder}

The purpose of serializing non-empty voxels is to find a path that traverses all input voxels. To preserve spatial topology, the serialization of 3D voxels can be viewed as a space-filling curve process. Existing space-filling curves mainly include Hilbert~\cite{Hilbert} and Z-order ~\cite{z_order} curves. While the Hilbert curve offers a better locality, constructing its index is computationally expensive. Thanks to the locality enhancement design (see Sec.~\ref{slm}), we do not heavily rely on preserving spatial topology information during serialization. Therefore, we opt for the computationally efficient Z-order curve. Specifically, given the coordinates of the input voxel set $\mathcal{C} = \{(x_{i}, y_{i}, z_{i}), i \in (1, N)\}$, where N denotes the number of voxels, the corresponding Z-order code $\mathcal{Z}^{P} = \{z_{i}^{P}, i \in (1, N) \}$ is computed by the bit-interleaving function~\cite{z_order}. Finally, sorting the Z-order code $\mathcal{Z}^{P}$ in ascending order to gain the Z-order index $\mathcal{Z}^{index} = \{z_{i}^{index}, i \in (1, N)\}$ for each voxel. 

Traditional Z-order curve using the X-axis as the primary order to calculate the index. However, we observe that indexing solely in the X-direction results in the preservation of spatial proximity only along the X-axis. Therefore, we introduce a variant that uses the Y-axis as the primary order for indexing. 

\subsubsection{Local-Global Sequential Aggregator}
\label{lgme}
Richer global contextual information is essential for enhancing 3D detection capabilities. Utilizing Mamba's computational advantages, treating the entire scene as a group to extract long-range global dependencies is intuitive. But the fine-grained local information is also crucial.
Therefore, we propose the Local-Global Sequential Aggregator (LGSA), which models the ``local and global" inter-dependencies simultaneously, as shown in Fig.~\ref{fig:lgsa}.

\noindent
\textbf{Global Sequential Encoder (GSE)} treats all non-empty voxels in the scene as a 1D sequence to directly establish a global receptive field. Unlike traditional window-based serialization, it does not create explicit local windows. The basic GSE layer includes two cascaded Mamba layers to encode the serialized voxels using Complementary Z-order. 
Specifically, given the input voxel sets $\mathcal{F} \in \mathbb{R}^{N \times C}$ and their corresponding voxel coordinates sets $\mathcal{C} \in \mathbb{R}^{N \times 3}$, We first serialize $\mathcal{F}$ using X-axis Z-order to obtain the corresponding voxel sequence $\mathcal{F}^{X}$ and coordinates $\mathcal{C}^{X}$. Subsequently, the reordered sequence $\mathcal{F}^{X}$ is input into the first Mamba layer to obtain the output $\mathcal{F}^{X}_{out}$. After that, the sequence $\mathcal{F}_{out}^{X}$ is reordered using the Y-axis Z-order curve to obtain the voxel sequence $\mathcal{F}^{Y}$ and coordinates $\mathcal{C}^{Y}$ and input into the second Mamba layer to produce the output $\mathcal{F}^{G}$ of the GSE. The specific calculations can be summarized as follows:

\begin{equation}
\begin{aligned}
     \mathcal{F}^{X}, \mathcal{C}^{X} &= {\rm Z\_INDEX\_X}(\mathcal{F}, \mathcal{C}), \\
     \mathcal{F}^{X}_{out} &= {\rm MAMBA_{1}}(\mathcal{F}^{X}),\\
     \mathcal{F}^{Y}, \mathcal{C}^{Y} &= {\rm Z\_INDEX\_Y}(\mathcal{F}^{X}_{out}, \mathcal{C}^{X}), \\
     \mathcal{F}^{G}&= {\rm MAMBA_{2}}(\mathcal{F}^{Y})
     \end{aligned}
\end{equation}
where In this context, $\rm Z\_INDEX\_X(\cdot)$ and $\rm Z\_INDEX\_Y(\cdot)$ represent the Z-order curve indexing functions for the X-axis and Y-axis, respectively. $\rm MAMBA_{i}(\cdot)$ refers to the $i_{th}$ basic Mamba layer in the GSE, where the input dimension is $C$. $\mathcal{F}^{G}_{out} \in \mathbb{R}^{N \times C}$ is the output of the GSE.

\noindent
\textbf{Local Sequential Encoder (LSE).} Unlike GSE, LSE is a window-based grouping architecture. Following the traditional window partitioning method~\cite{fan2022embracing,wang2023dsvt} based on transformers, We divide all the voxels into non-overlapping 3D windows. However, unlike the previous sequential traversal of windows, we apply the Z-order curve for traversal within the local windows. Specifically, given a window size of $(w_{x}, w_y, w_z)$, for each voxel $v_{i}$, we can compute its window coordinates $c_{win}=(\left \lfloor x_{i}/w_{x} \right \rfloor, \left \lfloor y_{i}/w_{y} \right \rfloor, \left \lfloor z_{i}/w_{z} \right \rfloor)$ and its local coordinates within window $c_{local}=(x_{i}-\left \lfloor x/w_{x} \right \rfloor \cdot w_{x},y_{i}-\left \lfloor y/w_{y} \right \rfloor \cdot w_{y},z_{i}-\left \lfloor z/w_{z} \right \rfloor \cdot w_{z})$. Then, we use the local window coordinates $c_{local}$ as input for the Z-order to obtain the voxel index within the local window. To enhance efficiency, we adopt the grouping method consistent with FlatFormer~\cite{liu2023flatformer}. After windows-based grouping, we partition $\mathcal{F}$ into multiple equal-length 1D sequences, denoted as $\mathcal{F}_{local} = \{f_{i},i\in [1, \left \lceil N/L \right \rceil ]\}$, where $L$ denotes the group size. The calculation of LSE is as follows:
\vspace{-0.1cm}
\begin{equation}
\begin{aligned}
     f^{X}_{ilocal}, c^{X}_{ilocal} &= {\rm Z\_INDEX\_X}(f_{local}, c_{local}), \\
     f^{X}_{ilout} &= {\rm MAMBA_{1}}(f^{X}_{ilocal}),\\
     f^{Y}_{ilocal}, c^{Y}_{ilocal} &= {\rm Z\_INDEX\_Y}(f^{X}_{ilout}, c^{X}_{ilocal}), \\
     f^{L}_{iout} &= {\rm MAMBA_{2}}(f^{Y}_{ilocal}),\\
     F^{L} &= Concat(f_{1out}^{L},\dots,f_{\left \lceil N/L \right \rceil out}^{L})
     \end{aligned}
\end{equation}

\begin{table*}[]
\centering
\caption{Comparison experiments on the nuScenes without using any test-time augmentation and model ensemble strategies. `C.V.', `Ped.', `Motor.', `B.C.', and `T.C.' represent construction vehicle, pedestrian, motorcycle, bicycle and traffic cone, respectively.}
\vspace{-0.2cm}
\scalebox{0.9}{
\begin{tabular}{lccccccccccccc}
\hline
\multicolumn{1}{l|}{Method}         & \multicolumn{1}{c|}{Present at} & mAP           & \multicolumn{1}{c|}{NDS}           & Car           & Truck         & Bus           & Trailer       & C.V.          & Ped.          & Motor.        & B.C.          & T.C.          & Barrier       \\ \hline
\multicolumn{14}{c}{Results on the val set}                                                                                                                                                                                                                                                \\ \hline
\multicolumn{1}{l|}{CenterPoint~\cite{yin2021center}}    & \multicolumn{1}{c|}{CVPR'2021}  & 59.2          & \multicolumn{1}{c|}{66.5}          & 84.9          & 57.4          & 70.7          & 38.1          & 16.9          & 85.1          & 59.0          & 42.0          & 69.8          & 68.3          \\
\multicolumn{1}{l|}{Transfusion-L~\cite{bai2022transfusion}}  & \multicolumn{1}{c|}{CVPR'2022}  & 65.5          & \multicolumn{1}{c|}{70.1}          & 86.9          & 60.8          & 73.1          & 43.4          & 25.2          & 87.5          & 72.9          & 57.3          & 77.2          & 70.3          \\
\multicolumn{1}{l|}{VoxelNeXt~\cite{chen2023voxelnext}}      & \multicolumn{1}{c|}{CVPR'2023}  & 60.5          & \multicolumn{1}{c|}{66.7}          & 83.9          & 55.5          & 70.5          & 38.1          & 21.1          & 84.6          & 62.8          & 50.0          & 69.4          & 69.4          \\
\multicolumn{1}{l|}{SAFDNet~\cite{zhang2024safdnet}}        & \multicolumn{1}{c|}{CVPR'2024}  & 66.3          & \multicolumn{1}{c|}{71.0}          & 87.6          & 60.8 & 78.0          & 43.5          & 26.6 & 87.8          & 75.5          & 58.0          & 75.0          & 69.7          \\ 
\multicolumn{1}{l|}{SEED~\cite{liu2024seed}}     & \multicolumn{1}{c|}{ECCV'2024}  & 66.2          & \multicolumn{1}{c|}{71.2}          & -          & -          & -          & -          & -          & -          & -          & -          & -          & -          \\ 
\multicolumn{1}{l|}{LION-Mamba~\cite{liu2024lion}}        & \multicolumn{1}{c|}{NIPS'2024}  & 68.0          & \multicolumn{1}{c|}{72.1}          & 87.9          & \textbf{64.9} & 77.6          & 44.4          & 28.5 & 89.6          & \textbf{75.6}         & \textbf{59.4}          & \textbf{80.8}          & 71.6          \\ 
\hline
\rowcolor{cyan!20} 
\multicolumn{1}{l|}{UniMamba (Ours)} & \multicolumn{1}{c|}{-}          & \textbf{68.5} & \multicolumn{1}{c|}{\textbf{72.6}} & \textbf{88.7} & 64.7          & \textbf{79.7} & \textbf{47.9} & \textbf{28.7}        & \textbf{89.7} & 74.6 & 59.1 & 79.5 & \textbf{72.3} \\ \hline
\multicolumn{14}{c}{Results on the test set}  
\\ \hline
\multicolumn{1}{l|}{3DSSD~\cite{yang20203dssd}}          & \multicolumn{1}{c|}{CVPR'2020}  & 42.6          & \multicolumn{1}{c|}{56.4}          & 81.2          & 47.2          & 61.4          & 30.5          & 12.6          & 70.2          & 36.0          & 8.6           & 31.1          & 47.9          \\
\multicolumn{1}{l|}{PillarNet~\cite{shi2022pillarnet}}      & \multicolumn{1}{c|}{ECCV'2022}  & 66.0          & \multicolumn{1}{c|}{71.4}          & 87.6          & 57.5          & 63.6          & 63.1          & 27.9          & 87.3          & 70.1          & 42.3          & 83.3          & 77.2          \\
\multicolumn{1}{l|}{Focals Conv~\cite{chen2022focal}}    & \multicolumn{1}{c|}{CVPR'2022}  & 63.8          & \multicolumn{1}{c|}{70.0}          & 86.7          & 56.3          & 67.7          & 59.5          & 23.8          & 87.5          & 64.5          & 36.3          & 81.4          & 74.1          \\
\multicolumn{1}{l|}{LargeKernel3D~\cite{chen2023largekernel3d}}  & \multicolumn{1}{c|}{CVPR'2023}  & 65.4          & \multicolumn{1}{c|}{70.6}          & 85.5          & 53.8          & 64.4          & 59.5          & 29.7          & 85.9          & 72.7          & 46.8          & 79.9          & 75.5          \\
\multicolumn{1}{l|}{DSVT~\cite{wang2023dsvt}}           & \multicolumn{1}{c|}{CVPR'2023}  & 68.4          & \multicolumn{1}{c|}{72.7}          & 86.8          & 58.4          & 67.3          & 63.1          & \textbf{37.1}          & 88.0          & 73.0          & 47.2          & 84.9          & 78.4          \\
\multicolumn{1}{l|}{HEDNet~\cite{zhang2023hednet}}        & \multicolumn{1}{c|}{NIPS'2023}  & 67.7          & \multicolumn{1}{c|}{72.0}          & 87.1          & 56.5          & 70.0          & 63.5 & 33.6 & 87.9          & 70.4          & 44.8          & 85.1          & 78.1          \\ 
\multicolumn{1}{l|}{SAFDNet~\cite{zhang2024safdnet}}        & \multicolumn{1}{c|}{CVPR'2024}  & 68.3     & \multicolumn{1}{c|}{72.3}          & 87.3          & 57.3          & 68.0          & 63.7 & 37.3 & 89.0          & 71.1          & 44.8          & 84.9          & \textbf{79.5}          \\ 
\rowcolor{cyan!20}    
\multicolumn{1}{l|}{UniMamba (Ours)} & \multicolumn{1}{c|}{-}          & \textbf{70.2} & \multicolumn{1}{c|}{\textbf{74.0}} & \textbf{87.9} & \textbf{60.4} & \textbf{70.9} & \textbf{65.9}          & 36.7          & \textbf{90.5} & \textbf{73.5} & \textbf{49.5} & \textbf{86.9} & 79.4 \\ \hline
\end{tabular}}
\label{nuscenes}
\end{table*}

\noindent
\textbf{Local-Global Aggregation.} To simultaneously extract local details and global context for each voxel, We adopt a channel grouping strategy to aggregate features with different receptive fields. For the input $\mathcal{F} \in \mathbb{R}^{N \times C}$, we divide it into M groups along the channel dimension, obtaining a set of voxel feature collections $\mathcal{F} = \{F_{i} \in \mathbb{R}^{N \times c} , i \in [1, M]\}$, where each group has a channel dimension of $c = C/M$. For the first $J$ groups, we apply GSEs for processing, while the remaining groups are processed using LSEs. Finally, we concatenate all the processed groups along the channel dimension to obtain the final output. The computation process can be expressed as follows:
\vspace{-0.1cm}
\begin{equation}
\begin{aligned}
    \mathcal{F}^{A} &= Concat[F_{1}^{G}, \dots, F_{J}^{G}, F_{J+1}^{L}, \dots, F_{M}^{L}],\\
    \mathcal{F}^{B} &= LayerNorm(\mathcal{F}^{A}) + \mathcal{F}^{A},\\
    \mathcal{F}^{A'} &= LayerNorm(FFN(\mathcal{F}^{B}) + \mathcal{F}^{B})
\end{aligned}
\end{equation}
Where $FFN(x) = Linear(Relu(Linear(x)))$ is a standard feedforward neural network, which is used to capture the feature interactions between different receptive fields.



\subsection{The UniMamba 3D Backbone}
\label{backbone}
With the proposed UniMamba Block, we build UniMamba 3D Backbone with flexible receptive fields and great spatial modeling capacity. The architecture of UniMamba is shown in Fig. \ref{fig:framework}. Following~\cite{zhang2023hednet}, we use an encoder-decoder architecture to stack UniMamba Blocks for further hierarchical feature extraction. 
As a general voxel-based 3D backbone, we replace the backbone in LION~\cite{liu2024lion} to construct our detector. For Argoverse 2~\cite{wilson2023argoverse}, we build on SAFDNet~\cite{zhang2024safdnet}. The configuration of the detection head and loss functions remains consistent with the baseline.


\section{Experiments}
\label{sec:exp}

\begin{table*}[]
\centering
\caption{Comparison experiments on the Waymo Open validation set. All of these models are trained with single-frame inputs and no additional test-time augmentation. ``Ped." denotes Pedestrian, and ``Cyc." denotes Cyclist.}
\vspace{-0.2cm}
\setlength{\tabcolsep}{0.22cm}
\scalebox{0.84}{
\begin{tabular}{l|c|c|c|c|c|c|c|c}
\hline
\multirow{2}{*}{Method} & \multirow{2}{*}{Present at}  & Vehicle(L1)          & Vehicle(L2)          & Ped.(L1)             & Ped.(L2)    & Cyc.(L1)             & Cyc.(L2) &ALL (L2)             \\
                        &                                                     & AP/APH             & AP/APH             & AP/APH             & AP/APH    & AP/APH             & AP/APH  &mAP/mAPH           \\ \hline
PointPillar~\cite{lang2019pointpillars}             & CVPR'2019                                       & 70.43/69.83          & 62.18/61.64          & 66.21/46.32          & 58.18/40.64 & 55.26/51.75          & 53.18/49.80  & 57.85/50.69       \\
SECOND~\cite{yan2018second}                  & Sensors'2018                                     & 70.96/70.34          & 62.58/62.02          & 65.23/54.24          & 57.22/47.49 & 57.13/55.62          & 54.97/53.53   &58.26/54.35       \\
CenterPoint~\cite{yin2021center}            & CVPR'2021                                       & 74.20/73.60          & 66.20/65.70          & 76.60/70.50          & 68.80/63.20 & 72.30/71.10                  & 69.70/68.50         &68.20/65.80         \\
Voxset~\cite{he2022voxelset}                 & CVPR'2022                                       & 74.50/74.00          & 66.00/65.60          & 80.00/72.40          & 72.50/65.40 & 71.60/70.30          & 69.00/67.70    &69.17/66.23      \\
SST~\cite{fan2022embracing}                                      & CVPR'2022                                     &75.13/74.64            &66.61/66.17           &80.07/72.12           &72.38/65.01  &71.49/70.20          &68.85/67.61    &69.28/66.26     \\
AFDetV2~\cite{hu2022afdetv2}                & AAAI'2022                                       & 77.64/77.14          & 69.68/69.22          & 80.19/72.62          & 72.16/66.95 & 73.72/72.74          & 71.06/70.12  &70.97/68.76        \\
PillarNet~\cite{shi2022pillarnet}             & ECCV'2022                                     & 79.09/78.59          & 70.92/70.46          & 80.59/74.01          & 72.28/66.17 & 72.29/71.21          & 69.72/68.67  &70.97/68.43        \\
PV\_RCNN++ ~\cite{shi2023pv}                                   &IJCV'2023                                        &79.25/78.78           &70.61/70.18          &81.83/76.28           &73.17/68.00   &73.72/72.66         &71.21/70.19 &71.66/69.46    \\  
OcTr ~\cite{zhou2023octr}                  & CVPR'2023                                      & 79.20/78.70          & 70.80/70.40          & 82.20/76.30          & 74.00/68.50 & 73.90/72.80          & 71.10/69.20   &71.97/69.37       \\
FlatFormer ~\cite{liu2023flatformer}  &CVPR'2023                               & - / -                    &69.00/68.60           & - / -               &71.50/65.30    & - / - &68.60/67.50   &69.70/67.13          \\ 
DSVT-V ~\cite{wang2023dsvt}  &CVPR'2023                               &79.70/79.30                    &71.40/71.00           &83.70/78.90               &76.10/71.50    &77.50/76.50 &74.60/73.70   &74.00/72.10          \\ 
MsSVT++ ~\cite{10371785}                                  & TPAMI'2023                                    &78.96/78.39           &70.57/70.01           &80.64/73.78           &73.12/66.80  &75.98/74.73           &72.97/71.82   &72.22/69.54       \\
HEDNet~\cite{zhang2023hednet} &NIPS'2023 &\textbf{81.10/80.60} &\textbf{73.20/72.70} &84.40/80.00 &76.80/72.60 &78.70/77.70 &75.80/74.90 &75.30/73.40 \\
SEED-B ~\cite{liu2024seed}  &ECCV'2024                             &79.70/79.20                    &71.80/71.40           &83.10/78.30               &75.50/70.80    &80.00/78.80 &77.30/76.10     &74.87/72.77        \\ 
VoxelMamba ~\cite{zhang2024voxel}  &NIPS'2024                             &80.80/80.30                    &72.60/72.20           &85.00/80.80               &77.70/73.60    &78.60/77.60 &75.70/74.80     &75.33/73.53        \\ 
LION-Mamba-L ~\cite{liu2024lion}  &NIPS'2024                             &80.30/79.90                    &72.00/71.60           &85.80/\textbf{81.40}               &78.50/\textbf{74.30}    &80.10/79.00 &77.20/76.20     &75.90/74.00        \\ 
\hline
\rowcolor{cyan!20} 
UniMamba (Ours)          & -    & 80.60/80.06 & 72.28/71.77 & \textbf{85.99}/81.25 & \textbf{78.66}/74.07 & \textbf{80.33/79.30} & \textbf{77.46/76.48} &\textbf{76.13/74.11} \\ 
\hline
\end{tabular}
}
\label{waymo}
\end{table*}

\begin{table*}[]
\centering
\caption{Comparison experiments on the Argoverse 2 val set. Here, C-Barrel, MPC-Sign, A-Bus, C-Cone, V-Trailer, MBT, W-Device, and W-Rider refer to Construction Barrel, Mobile Pedestrian Crossing Sign, Articulated Bus, Construction Cone, Vehicular Trailer, Message Board Trailer, Wheeled Device, and Wheeled Rider, respectively.}
\vspace{-0.2cm}
\setlength{\tabcolsep}{0.22cm}
\scalebox{0.55}{
\begin{tabular}{l|c|cccccccccccccccccccccccccc}
\hline
Method         & \rotatebox{90}{mAP}           & \rotatebox{90}{Vehicle} & \rotatebox{90}{Bus}  & \rotatebox{90}{Pedestrian} & \rotatebox{90}{Stop Sign} & \rotatebox{90}{Box Truck} & \rotatebox{90}{Bollard} & \rotatebox{90}{C-Barrel} & \rotatebox{90}{Motorcyclist} & \rotatebox{90}{MPC-Sign} & \rotatebox{90}{Motorcycle} & \rotatebox{90}{Bicycle} & \rotatebox{90}{A-Bus} & \rotatebox{90}{School Bus} & \rotatebox{90}{Truck} & \rotatebox{90}{Truck Cab} & \rotatebox{90}{C-Cone} & \rotatebox{90}{V-Trailer} & \rotatebox{90}{Sign} & \rotatebox{90}{Large Vehicle} & \rotatebox{90}{Stroller} & \rotatebox{90}{Bicyclist} & \rotatebox{90}{MBT} & \rotatebox{90}{Dog}  & \rotatebox{90}{Wheelchair} & \rotatebox{90}{W-Device} & \rotatebox{90}{W-Rider} \\ \hline
CenterPoint~\cite{yin2021center}    & 22.0          & 67.6 & 38.9 & 46.5 & 16.9 & 37.4     & 40.1 & 32.2 & 28.6      & 27.4 & 33.4       & 24.5    & 8.7   & 25.8 & 22.1  & 22.6  & 29.5 & 22.4 & 6.3  & 3.9  & 0.5  & 20.1     & 0.0 & 3.9  & 0.5   & 10.9  & 4.2   \\
FSDv1~\cite{fan2022fully}         & 28.2          & 68.1 & 40.9 & 59.0 & 29.0 & 38.5     & 41.8 & 42.6 & 39.7      & 26.2 & 49.0       & 38.6    & 20.4  & 30.5 & 21.1  & 14.8  & 41.2 & 26.9 & 11.9 & 5.9  & 13.8 & 33.4     & 0.0 & 9.5  & 7.1   & 14.0  & 9.2   \\
VoxelNeXt~\cite{chen2023voxelnext}     & 30.7          & 72.7 & 38.8 & 63.2 & 40.2 & 40.1     & 53.9 & 64.9 & 44.7      & 39.4 & 42.4       & 40.6    & 20.1  & 25.2 & 16.9  & 19.9  & 44.9 & 20.9 & 14.9 & 6.8  & 15.7 & 32.4     & 0.0 & 14.4 & 0.1   & 17.4  & 6.6   \\
HEDNet~\cite{zhang2023hednet}        & 37.1          & 78.2 & 47.7 & 67.6 & 46.4 & 45.9     & 56.9 & 67.0 & 48.7      & 46.5 & 58.2       & 47.5    & 23.3  & 40.9 & 21.6  & 27.5  & 46.8 & 27.9 & 20.6 & 6.9  & 27.2 & 38.7     & 0.0 & 30.7 & 9.5   & 28.5  & 8.7   \\
FSDv2~\cite{fan2023fsd}         & 37.6          & 77.0 & 47.6 & 70.5 & 43.6 & 41.5     & 53.9 & 58.5 & 56.8      & 39.0 & 60.7       & 49.4    & 28.4  & 41.9 & 24.0  & \textbf{30.2}  & 44.9 & \textbf{33.4} & 16.6 & 7.3  & 32.5 & 45.9     & \textbf{1.0} & 12.6 & \textbf{17.1}  & 26.3  & \textbf{17.2}  \\
SAFDNet~\cite{zhang2024safdnet}        & 39.7          & 78.5 & \textbf{49.4} & 70.7 & 51.5 & 44.7     & 65.7 & 72.3 & 54.3      & 49.7 & 60.8       & 50.0    & \textbf{31.3}  & \textbf{44.9} & 23.6  & 24.7  & 55.4 & 31.4 & 22.1 & 7.1  & 31.1 & 42.7     & 0.0 & 26.1 & 1.4   & 30.2  & 11.5  \\ 
LION-Mamba~\cite{liu2024lion}        & 41.5          & 75.1 & 43.6 & 73.9 & \textbf{53.9} & 45.1     & 66.4 & 74.7 & \textbf{61.3}      & 48.7 & \textbf{65.1}       & \textbf{56.2}    & 21.7  & 42.7 & 19.0  &25.3  & 58.4 & 28.9 & \textbf{23.6} & \textbf{8.3}  & \textbf{49.5} & 47.3     & 0.0 & \textbf{31.4} & 8.7   & \textbf{37.6}  & 11.8  \\ 
\hline
\rowcolor{cyan!20} 
UniMamba (Ours) & \textbf{42.0} &\textbf{78.9}      &47.9      &\textbf{74.3}      &51.8      &\textbf{46.8 }         &\textbf{67.8}      &\textbf{76.9}      &55.8           &\textbf{51.7}      &62.8   &52.2         &30.2       &44.6      &\textbf{24.6 }      &28.1       &\textbf{59.4}      &32.2      &23.2      &6.7      &41.5      &\textbf{48.5}          &0.0     &26.4      &8.1       &36.4       & 13.7      \\ \hline
\end{tabular}}
\label{av2}
\end{table*}


\subsection{Datasets and Metrics}

\noindent \textbf{nuScenes}~\cite{nuscenes} is a sophisticated outdoor dataset comprising 1,000 scenarios, with 700 allocated for training, 150 for validation, and 150 for testing. It encompasses a diverse array of object annotations across 10 categories, representing the majority of traffic participants. The size of objects varies significantly among these categories, which poses considerable challenges for detection. We evaluate the effectiveness of our approach using the provided assessment tools, and reporting metrics including NDS and mAP. 

\noindent \textbf{Waymo Open Dataset (WOD)}~\cite{sun2020scalability} is a large-scale dataset comprising 798 training scenarios and 202 validation scenarios, with 160,000 and 40,000 samples, respectively. To assess detection performance, the dataset toolkit provides metrics that include mean Average Precision (mAP) and mAPH, where mAPH is the mean average precision weighted by heading. There are two levels of detection difficulty defined based on the density of points within the bounding box: LEVEL\_1 (L1) includes bounding boxes with more than five points, while LEVEL\_2 (L2) encompasses bounding boxes containing one to five points.

\noindent \textbf{Argoverse 2}~\cite{wilson2023argoverse} is a  highly challenging dataset. It has an ultra-long detection range ($200m \times 200m$) and includes annotations for 26 different types of objects, posing significant challenges for both the real-time performance and detection capabilities. The dataset comprises a total of 1,000 scene sequences, with 700 designated for training and 150 for testing. We follow previous methods~\cite{fan2023fsd} by using mean Average Precision (mAP) to evaluate performance.

\subsection{Implementation Details}
We implement our UniMamba based on LION~\cite{liu2024lion}. Following previous methods, we use voxel sizes of $(0.3m, 0.3m, 0.25m)$, $(0.32m, 0.32m, 0.1875m)$, and $(0.4m, 0.4m, 0.25m)$ for voxelization of the nuScenes, Waymo, and Argoverse 2 datasets, respectively, where the number of voxel feature channels is set to 128. The downsampling stride for encoder-decoder is set to $\{1, 2, 2\}$, with each scale containing 1 layer of UniMamba Block. For each UniMamba block, the number of channel groups $M$ is set to 4, with $J=2$ groups applying GSE and the remaining two applying LSE. The other training hyperparameters are configured consistently with \cite{liu2024lion}. All experiments are conducted using 8 Tesla A800 GPUs, utilizing the AdamW~\cite{loshchilov2017decoupled} optimizer. For more network details, please refer to the appendix.

\subsection{Comparison Experiments}
\noindent \textbf{Result on nuscenes}
As shown in Tab. \ref{nuscenes}, our UniMamba demonstrates a significant performance improvement on the NuScenes dataset, being the first lidar-based method to exceed 70 mAP on the test set. Compared to the traditional SpCNN-based state-of-the-art SAFDNet~\cite{zhang2024safdnet}, it achieves \textit{improvements of 2.2 mAP and 1.6 NDS}. Additionally, when compared to the transformer-based method DSVT~\cite{wang2023dsvt}, UniMamba shows enhancements of 1.8 mAP and 1.3 NDS. This validates the suitability of the Mamba architecture as a 3D backbone. Notably,  UniMamba has significantly improved detection performance for various object sizes, achieving a \textit{3.9 mAP increase for Truck (large target) and a 1.9 mAP improvement for Pedestrian (small target)}. This highlights the effectiveness of UniMamba's hybrid architecture design and flexible receptive field.


\noindent \textbf{Result on Waymo}
We then compare UniMamba with previous methods on the Waymo dataset, establishing a new state-of-the-art. As shown in Tab. \ref{waymo}, our method outperforms the previous best SpCNN-based method, HEDNet~\cite{zhang2023hednet}, by 0.83 in L2\_mAP. Notably, compared to the Transformer-based method DSVT~\cite{wang2023dsvt}, UniMamba has demonstrated significant advantages in small object, achieving \textit{improvements of +2.56 L2\_mAP for pedestrian and +2.86 L2\_mAP for cyclist}. This emphasizes the critical importance of local details for the perception of small targets.

\noindent \textbf{Result on Argoverse 2}
We also evaluated UniMamba on the Argoverse 2 dataset, which has a larger detection range, with results shown in Tab. \ref{av2}. UniMamba achieved state-of-the-art performance, reaching 42.0 mAP. Compared to the previous state-of-the-art SAFDNet~\cite{zhang2024safdnet}, our method achieved an improvement of 2.3 mAP, by simply replacing its 3D backbone. This demonstrates the powerful 3D feature extraction capability of UniMamba.

\subsection{Ablation Study}
\begin{table}[]
\centering
\caption{Ablation Study of Spatial Locality Modeling module cross Serializations. For a fair comparison, the latency is calculated as the time taken to serialize 20,000 input voxels.}
\vspace{-0.2cm}
\setlength{\tabcolsep}{0.23cm}
\scalebox{0.8}{
\begin{tabular}{l|cc|cc|c}
\hline
\multirow{2}{*}{Serialization Method} & \multicolumn{2}{c|}{w/o SLM} & \multicolumn{2}{c|}{w/ SLM} & Latency \\
                                      & mAP            & NDS            & mAP            & NDS           &         \\ \hline
Random                                & 66.2           & 70.9           & 67.4           & 71.5          &-         \\
Hilbert                               & 67.4           & 71.7           & 67.6           & 71.8          &15.8 ms         \\
Z-order                              & 67.2           & 71.5           & 67.6           & 71.7          &0.9 ms         \\
\rowcolor{cyan!20} 
Complementary Z-order                & 67.6           & 71.9           & \textbf{67.9}  & \textbf{72.0} &\textbf{0.9 ms}         \\ \hline
\end{tabular}
}
\label{tab:LEConv}
\end{table}

\begin{table}[]
\centering
\caption{Ablation Study of the Number of Channel Groups in LGSA. We use Transfusion-L~\cite{bai2022transfusion} as the baseline.}
\vspace{-0.2cm}
\setlength{\tabcolsep}{0.35cm}
\begin{tabular}{c|cc|cc}
\hline
\multirow{2}{*}{Channel Groups} & \multirow{2}{*}{GSE} & \multirow{2}{*}{LSE} & \multicolumn{2}{c}{\multirow{2}{*}{mAP NDS}} \\
                                  &                      &                      & \multicolumn{2}{c}{}                         \\ \hline
1 (Baseline)                                & 0                    & 0                    &65.7                       &70.5                      \\
1 (LSE only)                                & 0                    & 1                    &67.7                       &72.0                      \\
1 (GSE only)                               & 1                    & 0                    &67.8                       &72.2                      \\ \hline
2                                 & 1                    & 1                    &68.4                       &72.5                      \\
\rowcolor{cyan!20} 
4                                 & 2                    & 2                    & \textbf{68.5}         & \textbf{72.6}        \\
4                                 & 1                    & 3                    &68.4                       &72.5                      \\
4                                 & 3                    & 1                    &68.3          &72.2             \\
8                                 & 4                    & 4                    &68.2          &72.3                      \\ \hline
\end{tabular}
\label{tab:LGME}
\vspace{-0.2cm}
\end{table}

\noindent
\textbf{Effect of Spatial Locality Modeling.}
To validate the effectiveness of the proposed SLM in mitigating the loss of spatial locality, we compare it with several popular space-filling curves. The "Random" refers to the configuration where no spatial curve is applied. 
As shown in rows 2-4 of Tab. \ref{tab:LEConv}, when neither SLM nor any spatial curve is used, performance drops significantly (- 1.4 mAP), demonstrating the importance of spatial locality. Compared with the standard Z-order curve, the Hilbert curve which better preserves locality shows a 0.2 mAP advantage. With SLM, the "Random" achieves 67.4 mAP, comparable to the performance with space-filling curves. Additionally, the Z-order curve achieves performance parity with the Hilbert curve (67.6 mAP) but requires only 1/17 time cost. These results demonstrate the effectiveness of Spatial Locality Modeling in capturing the dynamic structure embedding.

\noindent
\textbf{Effect of Complementary Z-order Serialization.}
As shown in row 5 of Tab. \ref{tab:LEConv}, Compared to the single-direction Z-order curve, using the Complementary Z-order curve brings a 0.3 mAP improvement. This demonstrates the effectiveness of using z-order for serialization from two directions to further enhance locality.

\noindent
\textbf{Effect of Design Choices in LGSA.}
We study the effect of LSE and GSE proposed in LGSA compared with the SpCNN-based method~\cite{bai2022transfusion}. Experimental results are shown in rows 2-4 of Tab.~\ref{tab:LGME}, which indicate that the SSM-based architecture offers significant advantages, surpassing the baseline by 2.0 mAP. This confirms the effectiveness of our proposed LSE and GSE.

\noindent
\textbf{Number of Channel Groups in LGSA.}
As shown in rows 4-8 of Tab. \ref{tab:LGME}, we conduct an ablation study on the number of channel groups and the quantities of LSE and GSE. It can be observed that using only two channel groups leads to a significant performance improvement (+0.6 mAP), which demonstrates the importance of simultaneously modeling both local and global receptive fields. The performance is optimal when the number of groups increases to 4. However, when using an imbalanced LSE and GSE, there is a decline in performance, indicating that LSE and GSE hold equal importance for subsequent detection tasks.

\noindent
\textbf{Different Local-Global Feature Aggregation Methods.}
To validate the effectiveness of the channel grouping strategy in aggregating local-global contextual information, we compare it with both sequential $``LSE-GSE"$ and parallel $``LSE\oplus GSE"$ approaches without channel grouping. As shown in Tab. \ref{tab:LGME2}, the channel grouping strategy achieves the best performance without introducing additional time overhead. This demonstrates the effectiveness of adaptively capturing multiple receptive fields between channel groups.


\begin{table}[]
\centering
\caption{Different Local-Global Feature Aggregation Methods.}
\vspace{-0.2cm}
\setlength{\tabcolsep}{0.3cm}
\begin{tabular}{c|ccc}
\hline
\multirow{2}{*}{Method} & \multirow{2}{*}{mAP} & \multirow{2}{*}{NDS}  & \multirow{2}{*}{Latency} \\
                        &                      &                                               &                          \\ \hline
Sequential              & 68.0                 & 72.3                                          &158.9ms                          \\
Parallel                & 68.4                 & 72.5                                          &162.7ms                          \\
\rowcolor{cyan!20} 
Channel Group (Ours)           & \textbf{68.5 }                & \textbf{72.6 }                                       &\textbf{121.4ms}                          \\ \hline
\end{tabular}
\label{tab:LGME2}
\end{table}

\begin{table}[]
\centering
\caption{Comparison of computational costs and parameters of different backbone architectures}
\vspace{-0.2cm}
\scalebox{0.8}{
\begin{tabular}{l|c|c|cc}
\hline
\multirow{2}{*}{Method} & \multirow{2}{*}{Backbone}    & \multirow{2}{*}{L2 mAP/mAPH} & \multirow{2}{*}{Flops} & \multirow{2}{*}{Params} \\
                        &                              &                              &                        &                         \\ \hline
CenterPoint             & SpCNN                        & 68.20/65.80                  & 48.5G                  & 2.7M                    \\ \hline
FlatFormer              & \multirow{2}{*}{Transformer} & 69.70/67.13                  & \textbf{48.5G}         & \textbf{1.1M}           \\
DSVT-V                  &                              & 74.00/72.10                  & 110.2G                 & 2.7M                    \\ \hline
\rowcolor{cyan!20}
UniMamba                & Mamba                        & \textbf{75.40/73.61}         & 61.9G                  & 1.6M                    \\ \hline
\end{tabular}
}
\label{tab:flop}
\vspace{-0.1cm}
\end{table}

\noindent
\textbf{Efficiency Analysis.}
To validate the computational efficiency of UniMamba, we compare it with mainstream 3D backbones based on SpCNN and Transformer. As shown in Tab. \ref{tab:flop}, UniMamba achieved an L2\_mAP performance of 75.40 with a computational load of 61.9 GFlops. Compared to DSVT, UniMamba improves the L2\_mAP by 1.4 while operating at approximately half the computational cost. These results demonstrate that UniMamba maintains strong detection accuracy with relatively low computational costs and parameter sizes.

\section{Conclusion}
\label{sec:conclusion}
In this paper, we propose a novel unified Mamba architecture, UniMamba, to integrate 3D convolution and State Space Models (SSM) in a concise multi-head format. UniMamba offers flexible receptive fields and excellent spatial modeling capabilities. We first analyze the challenges of directly applying the Mamba architecture in Lidar-based 3D detection tasks, including locality loss and limited spatial diversity, which restrict its ability to capture complex local and global dependencies. Then, we introduce a spatial locality modeling module to capture dynamic structural embeddings and use complementary Z-order curves to preserve spatial proximity when transforming 3D voxels into a 1D sequence. Our core contribution is the Local-Global Sequential Aggregator (LSGA), which includes a Local Sequential Encoder and Global Sequential Encoder to capture local and global receptive fields. Additionally, we adopt a channel-grouping strategy to efficiently aggregate these various receptive fields. Experiments on three challenging datasets (nuScenes, Waymo and Argoverse 2) demonstrate the superior performance of our UniMamba.

{
    \small
    \bibliographystyle{ieeenat_fullname}
    \bibliography{main}
}

\clearpage
\setcounter{page}{1}
\setcounter{figure}{0}
\setcounter{table}{0}
\maketitlesupplementary
\appendix

\label{sec:Suppl}
In the supplementary materials, we first provide more details on the network implementation (Sec. \ref{imp}), followed by additional ablation experiments (Sec. \ref{exp}), and finally present the qualitative results (Sec. \ref{vis}).

\section{More Implementation Details}
\label{imp}
\quad For the nuScenes \cite{nuscenes}, we first voxelize the point cloud with a range of $[-54m, -54m, -5m, 54m, 54m, 3m]$ using a voxel size of $[0.3m, 0.3m, 0.25m]$, resulting in a voxel feature of $[360, 360, 32]$. We then employ a 5-stage UniMamba 3D Backbone. Each stage has a downsampling stride of $\{1, 2, 2\}$, with $M =4$ groups in the UniMamba block and $J=2$ LSEs. The window size for the X/Y-axis in the LSE of each stage is $[13, 13]$, while the Z-axis window size varies across stages as $\{32, 16, 8, 4, 2\}$ (i.e., no subdivision of the window in the z-axis direction). To improve efficiency, we follow the Flatformer~\cite{liu2023flatformer} approach and group the windows in an equal-length manner, with a group size of 1024. The window size in the X/Y-axis for the GSE is the size of the entire space (i.e., $[360, 360]$), and the Z-axis window size is consistent with that of the LSE. The input and output dimensions of the backbone are set to $C=128$. The BEV backbone and detection head are consistent with DSVT~\cite{wang2023dsvt}.

For the Waymo Open Dataset~\cite{sun2020scalability} and Argoverse 2 dataset~\cite{wilson2023argoverse}, we voxelize point clouds using sizes of $[0.32m, 0.32m, 0.1875m]$ and $[0.4m, 0.4m, 0.25m]$ within ranges of $[-75.52m, -75.52m, -2m, 75.52m, 75.52m, 4m]$ and $[-200m, -200m, -4m, 200m, 200m, -4m]$, resulting in voxel features of $[472, 472, 32]$ and $[1000, 1000, 32]$, respectively. The configurations of the UniMamba 3D Backbone remain consistent with those of nuScenes. For a fair comparison, we use a sparse detection head consistent with SAFDNet~\cite{zhang2024safdnet} for the Argoverse 2 dataset. 
\section{More Ablation Experiments}
\label{exp}
\subsection{Different Window and Group Size in LSE}
\quad To verify the impact of different window shapes and group sizes on performance in LSE, we conduct ablation experiments on the nuScenes dataset. The results for different window shapes are shown in rows 2-4 of Tab. \ref{tab:lse}, where it can be observed that the performance of UniMamba is insensitive to window shape. The results for different group sizes are presented in rows 5-8 of Tab. \ref{tab:lse}, indicating that both smaller and larger group sizes lead to a decline in performance, with the best performance achieved at a group size of 1024. This is because a group that is too small results in a not enough receptive field, while a group that is too large lacks detailed local information.

\begin{table}[hb]
\centering
\caption{Comparisons of different window shapes and group sizes in LSE on the nuScenes \textit{val} set. The ``Window Shape" refers to the size of the window in the X/Y-axis, while the window size along the Z-axis is set to $\{32, 16, 8, 4, 2\}$ for different stages as default (\textit{i.e.}, no subdivision of the window in the Z-axis direction).}
\setlength{\tabcolsep}{0.28cm}
\begin{tabular}{c|c|cc}
\hline
\textbf{Window Shape}      &  \textbf{Group Size}       &  \textbf{mAP} $\uparrow$           & \textbf{NDS} $\uparrow$           \\ \hline
{[}9,9{]}                    & \multirow{3}{*}{1024} & 68.4          & 72.5          \\
{[}13,13{]}                  &                       & \textbf{68.5} & \textbf{72.6} \\
{[}18,18{]}                  &                       & 68.5          & 72.5          \\ \hline
\multirow{4}{*}{{[}13,13{]}} & 256                   & 68.3          & 72.4          \\
                             & 512                   & 68.4          & 72.6          \\
                             & 1024                  & \textbf{68.5} & \textbf{72.6} \\
                             & 2048                  & 68.4          & 72.4          \\ \hline
\end{tabular}
\label{tab:lse}
\end{table}

\subsection{Effect of Downsampling Strides}
\quad To verify the effectiveness of extracting hierarchical features using the encoder-decoder architecture like HEDNet\cite{zhang2023hednet}, we compare the performance of models with different downsampling strides. As shown in Tab. \ref{tab:MFE}, $\{1\}$ indicates the absence of the encoder-decoder architecture, while $\{1,2\}$ indicates only one downsampling. It can be observed that the mAP decreases sequentially for $\{1\}$ and $\{1,2\}$, which demonstrates that hierarchical features play a crucial role in detection tasks. Additionally, when the downsampling strides are $\{1,2,4\}$ and $\{1,4,4\}$, the performance also declines, indicating that excessively large downsampling strides will result in the loss of some local details.

\begin{table}[hb]
\centering
\caption{Ablation study of different scales in each stage.}
\vspace{-0.2cm}
\setlength{\tabcolsep}{0.6cm}
\begin{tabular}{c|cc}
\hline
\multirow{1}{*}{\textbf{Down Strides}} & \multirow{1}{*}{\textbf{mAP} $\uparrow$} &  \multirow{1}{*}{\textbf{NDS}$\uparrow$} \\ \hline
\{1\}                        &68.0                       &72.3                      \\
\{1,2\}                      &68.3                       &72.5                      \\
\{1,2,2\}                    &\textbf{68.5}                       &\textbf{72.6}                      \\
\{1,2,4\}                    & 68.1         &72.3         \\
\{1,4,4\}                    & 67.9  & 72.2 \\ \hline
\end{tabular}
\label{tab:MFE}
\end{table}

\begin{figure*}[]
\centering
\subfloat[SAFDNet]{
		\includegraphics[width=\linewidth]{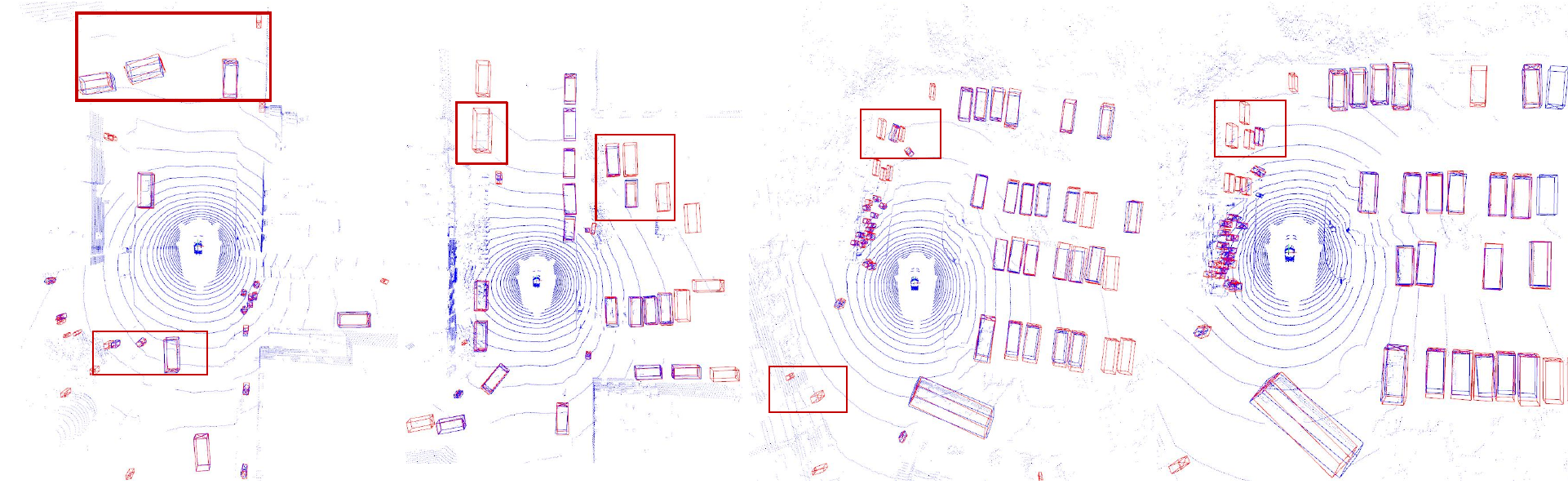}}
        \\
\subfloat[UniMamba]{
		\includegraphics[width=\linewidth]{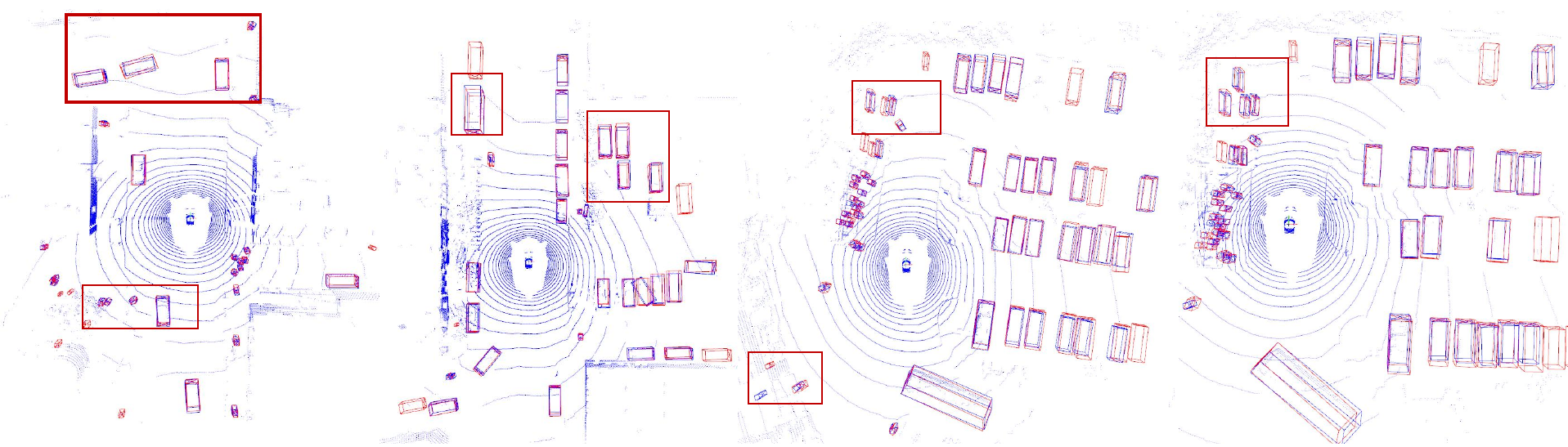}}
        \\
\caption{Visualization comparison of detection results between our UniMamba and SAFDNet~\cite{zhang2024safdnet} on the nuScenes \textit{val} set. \textcolor{blue}{Blue} indicates the prediction bounding box and \textcolor{red}{Red} indicates the ground-truth bounding box. The superior detection results are best viewed in \textcolor{red}{Red Rectangle}.}
\label{fig:vis}
\end{figure*}

\subsection{Complementary Order in Serialization}
\quad For the Complementary Z-order, we adopt two different feature interaction methods: parallel and Sequential. The parallel method refers to separately serializing and encoding the input features in the X-direction and Y-direction, and then summing them. In contrast, the Sequential method involves first serializing and encoding the input features in the X-direction, and then performing Y-direction serializing and encoding on the encoded features. The results are shown in Tab. \ref{z-order}, where the Sequential method yields the best performance.

\begin{table}[hb]
\centering
\caption{Comparison experiments of different interaction order in the proposed complementary serialization.}
\setlength{\tabcolsep}{0.7cm}
\begin{tabular}{c|cc}
\hline
\textbf{Method}     & \textbf{mAP} $\uparrow$ & \textbf{NDS} $\uparrow$ \\ \hline
Parallel   & 68.0 & 72.2 \\
Sequential & \textbf{68.5}            & \textbf{72.6}            \\ \hline
\end{tabular}
\label{z-order}
\end{table}

\section{Visualization Results}
\label{vis}
\quad To evaluate the qualitative results of UniMamba, we conduct a visual comparison of detection results with the previous state-of-the-art method, SAFDNet \cite{zhang2024safdnet}. As shown in Fig. \ref{fig:vis}, UniMamba demonstrates superior performance. For instance, pedestrians that are not detected by SAFDNet in the first column are successfully detected by UniMamba, and UniMamba's localization performance for vehicles also outperform those of SAFDNet. These results confirm that the flexible receptive field of UniMamba is effective in enhancing the model's detection capabilities.

\end{document}